# Benchmarking Emergency Department Triage Prediction Models with Machine Learning and Large Public Electronic Health Records

Feng Xie[1#], Jun Zhou[2#], Jin Wee Lee[1], Mingrui Tan[2], Siqi Li[1], Logasan S/O Rajnthern[3], Marcel Lucas Chee[4], Bibhas Chakraborty[1,5,6], An-Kwok Ian Wong[7], Alon Dagan[8,9], Marcus Eng Hock Ong[1,10], Fei Gao[2^], Nan Liu[1,11,12^]*

[1] Centre for Quantitative Medicine and Programme in Health Services and Systems Research, Duke-NUS Medical School, Singapore, Singapore
[2] Institute of High Performance Computing, Agency for Science, Technology and Research (A*STAR), Singapore, Singapore
[3] School of Electrical and Electronic Engineering, Nanyang Technological University, Singapore, Singapore
[4] Faculty of Medicine, Nursing and Health Sciences, Monash University, Victoria, Australia
[5] Department of Statistics and Data Science, National University of Singapore, Singapore, Singapore
[6] Department of Biostatistics and Bioinformatics, Duke University, Durham, NC, USA
[7] Division of Pulmonary, Allergy, and Critical Care Medicine, Duke University, Durham, NC, USA
[8] Department of Emergency Medicine, Beth Israel Deaconess Medical Center, Harvard Medical School, Boston, MA, USA
[9] MIT Critical Data, Laboratory for Computational Physiology, Institute for Medical Engineering and Science, Massachusetts Institute of Technology, Cambridge, MA, USA
[10] Department of Emergency Medicine, Singapore General Hospital, Singapore, Singapore
[11] SingHealth AI Health Program, Singapore Health Services, Singapore, Singapore
[12] Institute of Data Science, National University of Singapore, Singapore, Singapore

[#]Joint first author
[^]Joint senior author

* Correspondence: Nan Liu, Centre for Quantitative Medicine, Duke-NUS Medical School, 8 College Road, Singapore 169857, Singapore. Phone: +65 6601 6503. Email: liu.nan@duke-nus.edu.sg

## Abstract

The demand for emergency department (ED) services is increasing across the globe, particularly during the current COVID-19 pandemic. Clinical triage and risk assessment have become increasingly challenging due to the shortage of medical resources and the strain on hospital infrastructure caused by the pandemic. As a result of the widespread use of electronic health records (EHRs), we now have access to a vast amount of clinical data, which allows us to develop predictive models and decision support systems to address these challenges. To date, however, there are no widely accepted benchmark ED triage prediction models based on large-scale public EHR data. An open-source benchmarking platform would streamline research workflows by eliminating cumbersome data preprocessing, and facilitate comparisons among different studies and methodologies. In this paper, based on the Medical Information Mart for Intensive Care IV Emergency Department (MIMIC-IV-ED) database, we developed a publicly available benchmark suite for ED triage predictive models and created a benchmark dataset that contains over 400,000 ED visits from 2011 to 2019. We introduced three ED-based outcomes (hospitalization, critical outcomes, and 72-hour ED reattendance) and implemented a variety of popular methodologies, ranging from machine learning methods to clinical scoring systems. We evaluated and compared the performance of these methods against benchmark tasks. Our codes are open-source, allowing anyone with MIMIC-IV-ED data access to perform the same steps in data processing, benchmark model building, and experiments. This study provides future researchers with insights, suggestions, and protocols for managing raw data and developing risk triaging tools for emergency care.

## 1. Introduction

Emergency Departments (ED) experience large volumes of patient flows and growing resource demands, particularly during the current COVID-19 pandemic[1]. This growth has caused ED crowding[2] and delays in care delivery[3], resulting in increased morbidity and mortality[4]. The ED triage models[5-9] provide opportunities for identifying high-risk patients and prioritizing limited medical resources. ED triage centers on risk stratification, which is a complex clinical judgment based on factors such as the patient's likely acute course, availability of medical resources, and local practices.[10]

The widespread use of Electronic Health Records (EHR) has led to the accumulation of large amounts of data, which can be used to develop predictive models to improve emergency care[11,12]. Based on a few large-scale EHR databases, such as Medical Information Mart for Intensive Care III (MIMIC-III)[13], eICU Collaborative Research

Database[14], and Amsterdam University Medical Centers Database (AmsterdamUMCdb)[15], several benchmarks have been established[16-18]. These benchmarks standardized the process of transforming raw EHR data into readily usable data to construct predictive models. They have provided clinicians and methodologists with easily accessible and high-quality medical data, accelerating research and validation efforts[19,20]. These non-proprietary databases and open-source pipelines make it possible to reproduce and improve clinical studies in ways that would otherwise not be possible[16]. While there are some publicly available benchmarks, most pertain to intensive care settings, and there are no widely accepted benchmarks related to the ED. An ED-based public benchmark would lower the entry barrier for new researchers, allowing them to focus their efforts on novel research.

Machine learning has seen tremendous advances in recent years, and it has gained increasing popularity in the realm of ED triage prediction models[21-28]. These prediction models involve machine learning, deep learning, interpretable machine learning, and others. However, we have found that researchers often develop an ad-hoc model for one clinical prediction task at a time, using only one dataset[21-26]. There is a lack of comparative studies among different methods and models to predict the same ED outcome, undermining the generalizability of any single model. Generally, existing prediction models are developed on retrospective data without prospective validation in real-world clinical settings. Hence, there remains a need for prospective, comparative studies on the accuracy, interpretability, and utility of risk models in the ED. Using an extensive public EHR database, we aimed to standardize data preprocessing and establish a comprehensive ED benchmark dataset alongside comparable risk triaging models for three ED-based tasks. It is expected to facilitate reproducibility and model comparison, and accelerate progress toward utilizing machine learning in future ED-based studies.

In this paper, we proposed a public benchmark suite for the ED using a large EHR dataset and introduced three ED-based outcomes: hospitalization, critical outcomes, and 72-hour ED reattendance. We implemented and compared several popular methods for these clinical prediction tasks. We used data from the publicly available MIMIC IV Emergency Department (MIMIC-IV-ED) database[29,30], which contains over 400,000 ED visit episodes from 2011 to 2019. Our code is open-source (https://github.com/nliulab/mimic4ed-benchmark) so that anyone with access to MIMIC-IV-ED can follow our data processing steps, create benchmarks, and reproduce our experiments. This study provides future researchers with insights, suggestions, and protocols to process the raw data and develop models for emergency care in an efficient and timely manner.

## 2. Methods

This section consists of three parts. First, we describe raw data processing, benchmark data generation, and cohort formation. Next, we introduce baseline models for benchmark tasks. Finally, we elaborate on the experimental setup and model performance evaluation.

## 2.1 Master data generation

We standardized terminologies as follows. Patients are referred to by their *subjects_id*. Each patient has one or more ED visits, identified by *stay_id* in *edstays.csv*. If there is an inpatient stay following an ED visit, this *stay_id* could be linked with an inpatient admission, identified by *hadm_id* in *edstays.csv*. *subjects_id* and *hadm_id* can also be traced back to the MIMIC-IV[31] database to follow the patient throughout inpatient or ICU stay and patients' future or past medical utilization, if needed. In the context of our tasks, we used *edstays.csv* as the root table and *stay_id* as the primary identifier. As a general rule, we have one *stay_id* for each prediction in our benchmark tasks. All raw tables were linked through *extract_master_dataset.ipynb*, illustrated in Figure 1. The linkage was based on the root table, *and* merged through different identifiers, including *stay_id* (ED), *subjects_id*, *hadm_id*, or *stay_id* (ICU). We extracted all high-level information and consolidated them into a master dataset (*master_dataset.csv*).

To construct the master dataset, we reviewed a number of existing literature[5,7,32-34] to identify relevant variables and outcomes. Moreover, we consulted clinicians and informaticians familiar with the raw data and ED operation to identify and confirm all ED-relevant variables. We excluded variables that were irrelevant, repeated, or largely absent. A list of high-level constructed variables is presented in Table 1, including patient history, variables collected at triage and before ED disposition, and primary ED-relevant outcomes. The final master dataset includes 448,972 ED visits by 216,877 unique patients.

**Table 1.** List of high-level constructed variables in the master dataset and their origins and categories.

| Category | Sub-category | Source table (omit *.csv* below) | Variable description | Variable name in the master dataset |
|---|---|---|---|---|
| Patient history | Past ED visits | *edstay* | ED visits in the past month,<br>ED visits in the past three months,<br>ED visit in the past year | *n_ed_30d, n_ed_90d, n_ed_365d* |
| | Past hospitalizations | *admissions* | Hospitalizations in the past month,<br>Hospitalizations in the past three months,<br>Hospitalizations in the past year | *n_hosp_30d,*<br>*n_hosp_90d,*<br>*n_ hosp_365d* |
| | Past ICU admissions | *Icustays* | ICU admissions in the past month, | *n_icu_30d, n_icu_90d, n_icu_365d* |
| | | | ICU admissions in the past three months,<br>ICU admissions in the past year | |
| | Comorbidities | *diagnoses_icd, d_icd_diagnoses* | Charlson Comorbidity Index (CCI, 17 variables), Elixhauser Comorbidity Index (ECI, 30 variables) | *cci_** (* represents 17 variables),<br>*eci_** (* represents 30 variables) |
| Information at the triage station | Demographics | *patients* | Age, Gender | *age, gender* |
| | Triage-vital signs | *triage* | Emergency Severity Index (ESI)<br>Vital signs collected at triage:<br>Temperature (Celsius),<br>Heart rate (bpm),<br>Oxygen saturation (%),<br>Systolic blood pressure (mmHg),<br>Diastolic blood pressure (mmHg),<br>Pain scale | *triage_acuity* (ESI),<br>*triage_temperature,*<br>*triage_heartrate,*<br>*triage_o2sat,*<br>*triage_sbp,*<br>*triage_dbp,*<br>*triage_pain* |
| | Triage-chief complaint | *triage* | Top 10 chief complaints identified in the ED | *chiefcom_** (* represents ten different chief complaints) |
| Information before ED disposition | ED vital signs | *vitalsigns* | Vital signs collected during ED stay (last measurement):<br>Temperature (Celsius),<br>Heart rate (bpm),<br>Oxygen saturation (%),<br>Systolic blood pressure (mmHg),<br>Diastolic blood pressure (mmHg) | *ed_temperature,*<br>*ed_heartrate,*<br>*ed_o2sat,*<br>*ed_sbp,*<br>*ed_dbp* |
| | ED administrative | *edstays* | ED length of stay (hours) | *ed_los* |
| | Medication reconciliation | *medrecon* | Counts of medication reconciliation | *n_medrecon* |
| | Medication prescription | *pyxis* | Counts of medication prescription in current ED stay | *n_med* |
| Outcomes | Hospitalization | *edstays:hadm_id* | Whether the patient is admitted to inpatient stay following the current ED visit | *outcome_ hospitalization* |
| | Inpatient mortality | *patients:dod admissions:dischtime* | Whether the patient dies in the hospital before discharge | *outcome_inhospital_mortality* |
| | ICU transfer from ED | *icustays:intime, edstays:outtime* | Whether the patient is transferred to ICU from ED within 12 hours | *outcome_icu_transfer_12h* |
| | ED reattendance | *edstays* | Whether the patient revisits ED after the discharge from the index | *outcome_ed_revisit_3d* |
| | | | ED visit within three days (72 hours or 3 days) | |
| | Critical outcomes | *master_dataset: outcome_icu_transfer_12h, outcome_inhospital _mortality* | Whether the patient fulfills either inpatient mortality or ICU transfer within 12 hours | *outcome_critical* |

* denotes the task-specific wildcard string

### 2.2 Data processing and benchmark dataset generation

The data processing workflow *(data_general_processing.ipynb)*, illustrated in Figure 2, begins with the master dataset generated from Section 2.1 to generate the benchmark dataset. In the first step, we filtered out all ED visits with patients under 18 years old and those without primary emergency triage class assignments. A total of 441,437 episodes remained after the filtering process.

The raw EHR data cannot be used directly for model building due to missing values, outliers, duplicates, or incorrect records caused by system errors or clerical mistakes. We addressed these issues with several procedures. For vital signs and lab tests, a value would be considered an outlier and marked as missing if it was outside the plausible physiological range as determined by domain knowledge, such as a value below zero or a $SpO_2$ level greater than 100%. We followed the outlier detection procedure used in MIMIC-EXTRACT[18], a well-known data processing pipeline for MIMIC-III. We utilized the thresholds available in the source code repository of Harutyunyan at el.[35], where one set of upper and lower thresholds was used for filtering outliers. Any value that falls outside of this range was marked as missing. Another set of thresholds was introduced to indicate the physiologically valid range, and any value that falls beyond this range was replaced by its nearest valid value. These thresholds were suggested by clinical experts based on domain knowledge.

For benchmarking purposes, we fixed a test set of 20% (n=88,287) of ED episodes, covering 65,169 unique patients. Future researchers are encouraged to use the same test set for model comparisons and to interact with the test set as infrequently as possible. The training set consisted of the remaining 80% of ED episodes. The validation set can be derived from the training set if needed. Missing values (including outliers marked as missing and those initially absent) were imputed. In this project, we used the median values from the training set and other options are provided through our code repository. The same values were used for imputation on the test set.

**Figure 1.** Raw data and the linkage through four unique identifiers (omit *.csv* for table name)

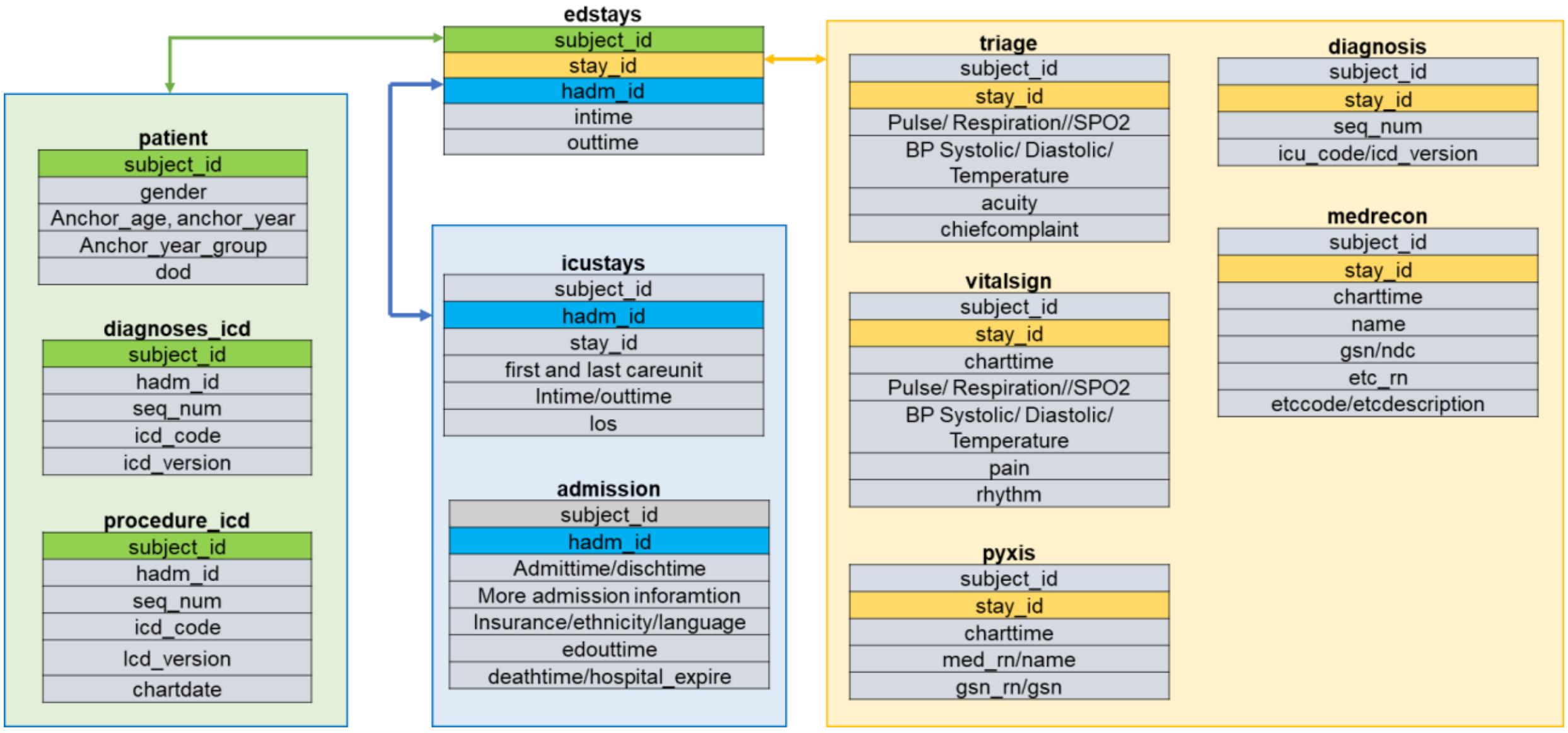

**Figure 2.** The workflow of data processing from raw data

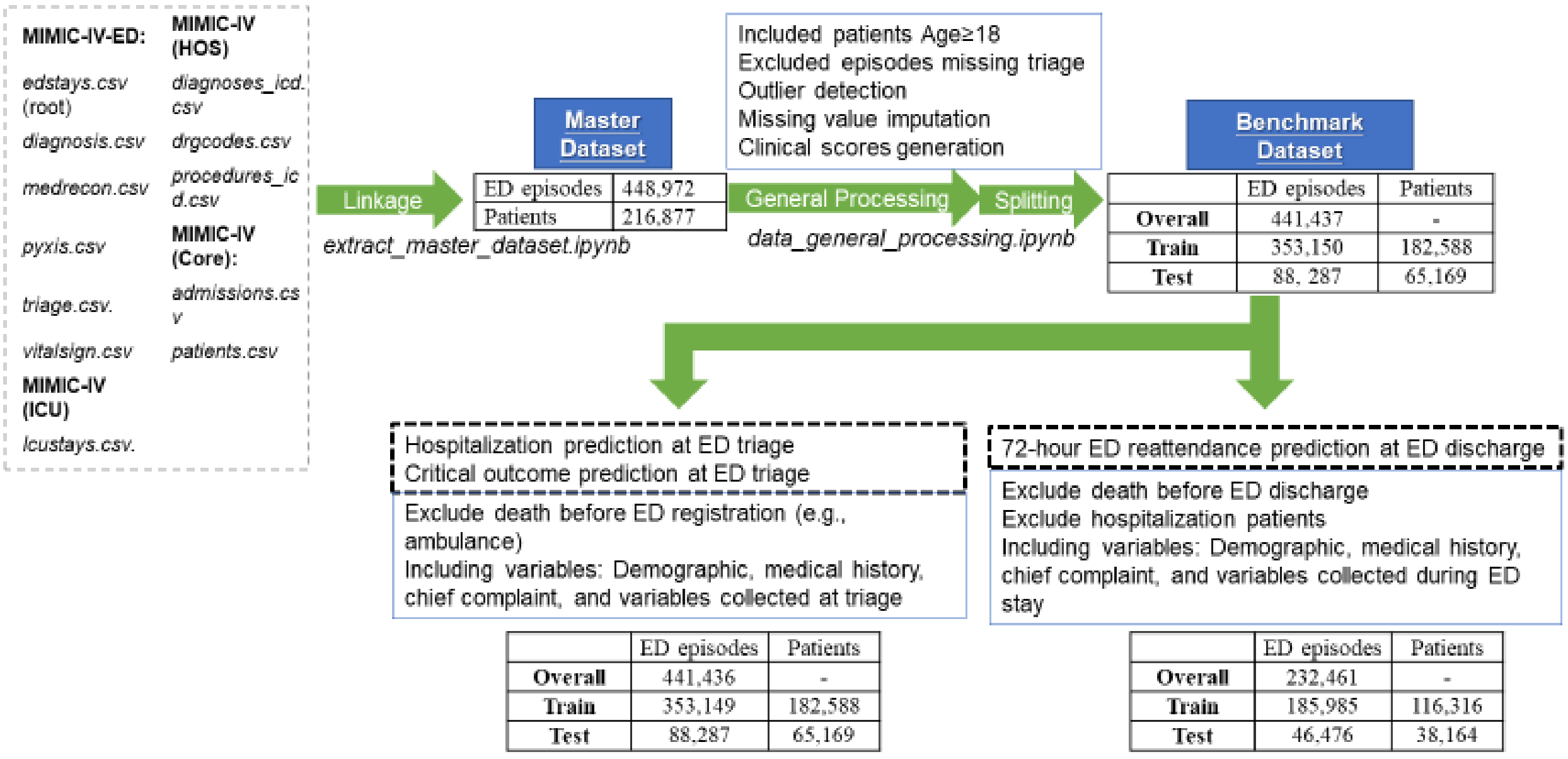

### 2.3 ICD codes processing

In MMIC-IV, each hospital admission is associated with a group of ICD diagnosis codes (in *diagnoses_icd.csv*), indicating the patients' comorbidities. We embedded the ICD codes within a time range (e.g., five years) from each ED visit into Charlson Comorbidity Index (CCI)[36] and Elixhauser Comorbidity Index (ECI)[37] according to the mapping proposed by Quan H et al.[38] We adopted the codebase from Cates et al.[39] and developed the neural network-based embedding with similar network structures to Med2Vec[40].

### 2.4 Benchmark tasks

Following are three ED-relevant clinical outcomes. They are all of utmost importance to clinicians and hospitals due to their immense implications on costs, resource prioritization, and patients' quality of life. Accurate prediction of these outcomes with the aid of big data and artificial intelligence has the potential to transform health services.

- The hospitalization outcome is met with an inpatient care site admission immediately following an ED visit[41-43]. Patients who transitioned to ED observation were not considered hospitalized unless they were eventually admitted to the hospital. As hospital beds are limited, this outcome indicates resource utilization and may facilitate resource allocation efforts. The hospitalization outcome also suggests patient acuity, albeit in a limited way, since hospitalized patients represent a broad spectrum of disease severity.
- The critical outcome[33] is compositely defined as either inpatient mortality[44] or transfer to an ICU within 12 hours. This outcome represents the critically ill patients who require ED resources urgently and may suffer from poorer health outcomes if care is delayed. Predicting the critical outcome at ED triage may enable physicians to allocate ED resources efficiently and intervene on high-risk patients promptly.
- The ED reattendance outcome refers to a patient's return visit to ED within 72 hours after their previous discharge from the ED. It is a widely used indicator of the quality of care and patient safety and is believed to represent patients who may not have been adequately triaged during their first emergency visit[45].

### 2.5 Baseline methods

Various triage systems, including clinical judgment, scoring systems, regression, machine learning, and deep learning, were applied to the benchmark dataset and evaluated on each benchmark task, as detailed in Table 2. A five-level triage system, Emergency Severity Index (ESI)[46], was assigned by a registered nurse based on clinical judgments. Level 1 is the highest priority, and level 5 is the lowest. Several scoring systems were also calculated, including the Modified Early Warning Score (MEWS)[47], National Early Warning Score (NEWS, versions 1 and 2)[48], Rapid

Emergency Medicine Score (REMS)[49], and Cardiac Arrest Risk Triage (CART)[50]. It is important to note that there are no neurological features (i.e., Glasgow Coma Scale) in the MIMIC-IV-ED dataset, which may lead to incomplete scores. Three machine learning methods – logistic regression (LR), random forest (RF), and gradient boosting (GB) – were benchmarked as well as deep learning methods multilayer perceptron (MLP)[51], Med2Vec[40], and long short-term memory (LSTM)[52-54]. These neural network structures are illustrated in eFigure 1. We used the scikit-learn package[55] with the default parameters for machine learning methods and Keras[56] for deep learning methods. In addition, the interpretable machine learning method, AutoScore[57,58], was implemented with its R software package[59].

**Table 2.** Description of various baseline methods

| | **Description** | **Variables** | **Hyperparameters** | **Package used** |
|---|---|---|---|---|
| Traditional machine learning | | | | |
| Logistic regression (LR) | Use the logistic function to model binary outcomes | Vitals, chief complaints, comorbidities, and age | penalty='l2', C=1.0, max_iter=100 | scikit-learn Python package |
| Random forest (RF) | Build many decision trees in parallel and combine the results through ensemble learning | | N_estimators=100 | |
| Gradient boosting (GB) | Build a number of decision trees in stages and combine the results along the way | | Loss='deviance', learning_rate=0.1, n_estimators=100 | |
| Traditional clinical scoring systems | | | | |
| Clinical Score: NEWS, NEWS2, MEWS, REMS, CART | Widely used clinical score for risk stratification at ED triage | Vitals, comorbidities, and age | None; No training is needed | None |
| Emergency Severity Index (ESI) | A subjective five-level triage system assigned by a registered nurse | *triage_acuity* | None | None |
| Interpretable machine learning | | | | |
| AutoScore | Interpretable machine learning automatic clinical score generator | Vitals, chief complaints, comorbidities, and age | Number of variables, tuned through performance-based parsimony plot | AutoScore R package |
| Deep learning | | | | |
| Multilayer perceptron (MLP) | The neural networks of multiple fully connected neurons | Vitals, chief complaints, comorbidities, and age | activation='relu', learning_rate=0.001, batch_size=200, epochs=20, loss=binary_crossentropy, optimizer = Adam | Keras Python package |
| Med2Vec | Embedding ICD codes with neural network | Vitals, chief complaints, comorbidities, age and ICD codes in the past 5 years | activation='relu', learning_rate=0.001, batch_size=200, epochs=100, loss=binary_crossentropy, optimizer = Adam | |
| LSTM | A special type of RNN which is capable of learning long-term dependencies | Basic static variables, and temporal variables of vital signs collected in the ED | activation='relu', learning_rate=0.001, batch_size=200, epochs=20, loss=binary_crossentropy, optimizer = Adam | |

CART: Cardiac Arrest Risk Triage
LSTM: Long Short-Term Memory
MEWS: Modified Early Warning Score
NEWS: National Early Warning Score
NEWS: National Early Warning Score, Version 2
REMS: Rapid Emergency Medicine Score
RNN: Recurrent Neural Network

### 2.6 Experiments, settings, and evaluation

We conducted all experiments on a server equipped with an Intel Xeon W-2275 processor, 128GB of memory, and an Nvidia RTX 3090 GPU, and the running time at model training was recorded. Deep learning models were trained using the Adam optimizer and binary cross-entropy loss. The AutoScore method optimized the number of variables through a parsimony plot. As the implementation was only for demonstration purposes, Module 5 of the clinical fine-tuning process in AutoScore was not implemented. We conducted the receiver operating characteristic (ROC) and precision-recall curve (PRC) analysis to evaluate the performance of all triage prediction models. The area under the ROC curve (AUROC) and the area under the PRC (AUPRC) values were reported as an overall measurement of predictive performance. Model performance was reported on the test set, and 100 bootstrapped samples were applied to calculate 95% confidence intervals (CI). Furthermore, we computed the sensitivity and specificity measures under the optimal cutoffs, defined as the points nearest to the upper-left corner of the ROC curves.

# 3. Results

## 3.1 Baseline characteristics of the benchmark dataset

We compiled a master dataset comprising 448,972 ED visits of 216,877 unique patients. After excluding incomplete or pediatric visits, a total of 441,437 adult ED visits were finally included in the benchmark dataset. They were randomly split into 80% (353,150) training data and 20% (88,287) test data. Table 3 summarizes the baseline characteristics of the entire cohort, stratified by outcomes. The average age of the patients was 52.8 years old, and 54.1% (n=242,844) of them were females. Compared with other patients, those with critical outcomes displayed a higher body temperature and heart rate and were prescribed a greater amount of medication. Additionally, they were more likely to have fluid and electrolyte disorders, coagulopathy, cancer, cardiac arrhythmias, valvular disease, and pulmonary circulation disorders.

**Table 3.** Characteristics of the benchmark dataset with a total of 81 included variables. Continuous variables are presented as *mean (SD)*; binary or categorical variables are presented as *count (%)*.

| | **Overall** | **Outcomes** | | | |
|---|---|---|---|---|---|
| | | **Hospitalization outcome** | | **Critical outcomes** | **72-hour ED reattendance** |
| | | **Discharge** | **Hospitalized** | | |
| # Emergency visits | 441,437 | 232,461 | 208,976 | 26,145 | 15,299 |
| *Demographic* | | | | | |
| Age | 52.80 (20.60) | 46.29 (19.36) | 60.03 (19.50) | 65.42 (17.85) | 50.40 (18.70) |
| Gender | | | | | |
| Female | 239794 (54.3%) | 133874 (57.6%) | 105920 (50.7%) | 12150 (46.5%) | 7068 (46.2%) |
| Male | 201643 (45.7%) | 98587 (42.4%) | 103056 (49.3%) | 13995 (53.5%) | 8231 (53.8%) |
| *Triage/scoring systems* | | | | | |
| Emergency Severity Index | | | | | |
| Level 1 | 25363 (5.7%) | 5349 (2.3%) | 20014 (9.6%) | 8874 (33.9%) | 462 (3.0%) |
| Level 2 | 147178 (33.3%) | 45445 (19.5%) | 101733 (48.7%) | 14087 (53.9%) | 3838 (25.1%) |
| Level 3 | 237565 (53.8%) | 151843 (65.3%) | 85722 (41.0%) | 3173 (12.1%) | 9849 (64.4%) |
| Level 4 | 30160 (6.8%) | 28704 (12.3%) | 1456 (0.7%) | 11 (0.0%) | 1091 (7.1%) |
| Level 5 | 1171 (0.3%) | 1120 (0.5%) | 51 (0.0%) | 0 (0.0%) | 59 (0.4%) |
| CART | 4.17 (5.06) | 2.68 (3.87) | 5.82 (5.67) | 8.66 (7.47) | 3.40 (4.28) |
| REMS | 3.56 (2.78) | 2.77 (2.60) | 4.43 (2.72) | 5.29 (2.63) | 3.20 (2.56) |
| NEWS | 0.91 (1.24) | 0.69 (0.95) | 1.16 (1.46) | 1.90 (2.10) | 0.91 (1.11) |
| NEWS2 | 0.80 (1.11) | 0.64 (0.90) | 0.98 (1.29) | 1.60 (1.82) | 0.80 (1.02) |
| MEWS | 1.36 (0.86) | 1.24 (0.71) | 1.49 (0.99) | 1.91 (1.34) | 1.35 (0.82) |
| *Previous health utilization* | | | | | |
| ED visit in the past month | 0.24 (0.79) | 0.21 (0.78) | 0.27 (0.79) | 0.20 (0.56) | 1.12 (2.32) |
| ED visit in the past 3 months | 0.53 (1.60) | 0.46 (1.59) | 0.62 (1.62) | 0.47 (1.07) | 2.36 (4.83) |
| ED visit in the past year | 1.42 (4.20) | 1.25 (4.16) | 1.61 (4.24) | 1.16 (2.65) | 6.06 (12.66) |
| Hospitalizations in the past month | 0.16 (0.52) | 0.09 (0.41) | 0.23 (0.60) | 0.21 (0.50) | 0.56 (1.31) |
| Hospitalizations in the past 3 months | 0.37 (1.03) | 0.21 (0.82) | 0.53 (1.20) | 0.50 (0.97) | 1.23 (2.76) |
| Hospitalizations in the past year | 0.98 (2.69) | 0.61 (2.20) | 1.39 (3.10) | 1.22 (2.33) | 3.37 (7.58) |
| ICU admissions in the past month | 0.02 (0.15) | 0.01 (0.10) | 0.03 (0.20) | 0.07 (0.30) | 0.02 (0.17) |
| ICU admissions in the past 3 months | 0.05 (0.26) | 0.02 (0.16) | 0.08 (0.34) | 0.17 (0.53) | 0.06 (0.30) |
| ICU admissions in the past year | 0.11 (0.49) | 0.05 (0.31) | 0.18 (0.63) | 0.37 (0.97) | 0.17 (0.61) |
| *Information collected at triage* | | | | | |
| Temperature (Celsius) | 36.71 (0.54) | 36.68 (0.49) | 36.75 (0.59) | 36.75 (0.66) | 36.69 (0.51) |
| Mean arterial pressure (mmHg) | 96.59 (14.86) | 97.55 (13.84) | 95.51 (15.86) | 92.08 (17.86) | 97.91 (14.77) |
| Heart rate (bpm) | 85.05 (17.46) | 83.90 (16.32) | 86.32 (18.56) | 90.73 (20.92) | 87.07 (16.94) |
| Respiratory rate (bpm) | 17.57 (2.49) | 17.30 (2.11) | 17.87 (2.83) | 18.91 (4.32) | 17.42 (2.16) |
| Oxygen saturations (%) | 98.40 (2.42) | 98.80 (2.00) | 97.95 (2.75) | 97.30 (3.70) | 98.39 (2.51) |
| Systolic blood pressure (mmHg) | 134.84 (22.14) | 135.14 (20.67) | 134.51 (23.67) | 129.18 (26.21) | 135.09 (21.79) |
| Diastolic blood pressure (mmHg) | 77.46 (14.71) | 78.76 (13.76) | 76.01 (15.57) | 73.53 (16.46) | 79.33 (14.62) |
| Pain scale | 4.15 (3.60) | 4.67 (3.58) | 3.58 (3.54) | 3.08 (3.02) | 4.74 (3.78) |
| *Chief complaints* | | | | | |
| Chest pain | 30756 (7.0%) | 13790 (5.9%) | 16966 (8.1%) | 1105 (4.2%) | 907 (5.9%) |
| Abdominal pain | 50868 (11.5%) | 25801 (11.1%) | 25067 (12.0%) | 1710 (6.5%) | 1961 (12.8%) |
| Headache | 16601 (3.8%) | 11967 (5.1%) | 4634 (2.2%) | 620 (2.4%) | 627 (4.1%) |
| Shortness of breath | 1285 (0.3%) | 402 (0.2%) | 883 (0.4%) | 213 (0.8%) | 24 (0.2%) |
| Back pain | 17625 (4.0%) | 12369 (5.3%) | 5256 (2.5%) | 282 (1.1%) | 621 (4.1%) |
| Cough | 9269 (2.1%) | 5293 (2.3%) | 3976 (1.9%) | 410 (1.6%) | 244 (1.6%) |
| Nausea/vomiting | 10666 (2.4%) | 5606 (2.4%) | 5060 (2.4%) | 466 (1.8%) | 401 (2.6%) |
| Fever/chills | 15267 (3.5%) | 4651 (2.0%) | 10616 (5.1%) | 1427 (5.5%) | 398 (2.6%) |
| Syncope | 8198 (1.9%) | 4409 (1.9%) | 3789 (1.8%) | 359 (1.4%) | 167 (1.1%) |
| Dizziness | 10928 (2.5%) | 6337 (2.7%) | 4591 (2.2%) | 365 (1.4%) | 287 (1.9%) |
| *Comorbidities (Charlson Comorbidity Index)* | | | | | |
| Myocardial infarction | 24773 (5.6%) | 6487 (2.8%) | 18286 (8.8%) | 2804 (10.7%) | 1080 (7.1%) |
| Congestive heart failure | 40784 (9.2%) | 10253 (4.4%) | 30531 (14.6%) | 5183 (19.8%) | 1285 (8.4%) |
| Peripheral vascular disease | 21985 (5.0%) | 5706 (2.5%) | 16279 (7.8%) | 2609 (10.0%) | 658 (4.3%) |
| Stroke | 21104 (4.8%) | 6431 (2.8%) | 14673 (7.0%) | 2385 (9.1%) | 745 (4.9%) |
| Dementia | 7387 (1.7%) | 2039 (0.9%) | 5348 (2.6%) | 887 (3.4%) | 252 (1.6%) |
| Chronic pulmonary disease | 62610 (14.2%) | 23142 (10.0%) | 39468 (18.9%) | 5354 (20.5%) | 3115 (20.4%) |
| Rheumatoid disease | 9115 (2.1%) | 3013 (1.3%) | 6102 (2.9%) | 774 (3.0%) | 273 (1.8%) |
| Peptic ulcer disease | 8315 (1.9%) | 2306 (1.0%) | 6009 (2.9%) | 899 (3.4%) | 318 (2.1%) |
| Liver disease | | | | | |
| None | 402913 (91.3%) | 220993 (95.0%) | 181920 (87.1%) | 22492 (86.0%) | 12695 (83.0%) |
| Mild liver disease | 29645 (6.7%) | 9489 (4.1 %) | 20156 (9.6 %) | 2581 (9.9 %) | 2153 (14.1%) |
| Moderate/severe liver disease | 8879 (2.0%) | 1979 (0.9 %) | 6900 (3.3 %) | 1072 (4.1 %) | 451 (2.9%) |
| Diabetes | | | | | |
| None | 355132 (80.5%) | 204810 (88.2%) | 150322 (72.0%) | 18020 (68.9%) | 11591 (75.8%) |
| Diabetes without chronic complications | 58375 (13.2%) | 19874 (8.5%) | 38501 (18.4%) | 5225 (20.0%) | 2649 (17.3%) |
| Diabetes with complications | 27930 (6.3%) | 7777 (3.3%) | 20153 (9.6%) | 2900 (11.1%) | 1059 (6.9%) |
| Hemiplegia | 5085 (1.2%) | 1573 (0.7%) | 3512 (1.7%) | 656 (2.5%) | 177 (1.2%) |
| Moderate to severe chronic kidney disease | 42952 (9.7%) | 11060 (4.8%) | 31892 (15.3%) | 4730 (18.1%) | 1263 (8.3%) |
| Cancer | | | | | |
| None | 401805 (91.0%) | 222186 (95.6%) | 179619 (85.9%) | 21561 (82.5%) | 14195 (92.8%) |
| Local tumor, leukemia, and lymphoma | 28631 (6.5%) | 7746 (3.3%) | 20885 (10.0%) | 3116 (11.9%) | 842 (5.5%) |
| Metastatic solid tumor | 11001 (2.5%) | 2529 (1.1%) | 8472 (4.1%) | 1468 (5.6%) | 262 (1.7%) |
| AIDS | 4079 (0.9%) | 1578 (0.7%) | 2501 (1.2%) | 258 (1.0%) | 426 (2.8%) |
| *Elixhauser Comorbidity Index* | | | | | |
| Cardiac arrhythmias | 61501 (13.9%) | 18815 (8.1%) | 42686 (20.4%) | 6590 (25.2%) | 2746 (17.9%) |
| Valvular disease | 22464 (5.1%) | 6210 (2.7%) | 16254 (7.8%) | 2646 (10.1%) | 702 (4.6%) |
| Pulmonary circulation disorders | 20357 (4.6%) | 5607 (2.4%) | 14750 (7.1%) | 2561 (9.8%) | 739 (4.8%) |
| Hypertension, uncomplicated | 44612 (10.1%) | 11542 (5.0%) | 33070 (15.8%) | 5018 (19.2%) | 1344 (8.8%) |
| Hypertension, complicated | 107846 (24.4%) | 39697 (17.1%) | 68149 (32.6%) | 8270 (31.6%) | 5214 (34.1%) |
| Other neurological disorders | 33515 (7.6%) | 11194 (4.8%) | 22321 (10.7%) | 3245 (12.4%) | 2292 (15.0%) |
| Hypothyroidism | 29407 | 9900 | 19507 (9.3%) | 2642 | 963 (6.3%) |
| | (6.7%) | (4.3%) | | (10.1%) | |
| Lymphoma | 4832 (1.1%) | 1253 (0.5%) | 3579 (1.7%) | 469 (1.8%) | 112 (0.7%) |
| Coagulopathy | 31206 (7.1%) | 8389 (3.6%) | 22817 (10.9%) | 3772 (14.4%) | 1597 (10.4%) |
| Obesity | 39138 (8.9%) | 14919 (6.4%) | 24219 (11.6%) | 2883 (11.0%) | 1525 (10.0%) |
| Weight loss | 23216 (5.3%) | 6448 (2.8%) | 16768 (8.0%) | 2607 (10.0%) | 1216 (7.9%) |
| Fluid and electrolyte disorders | 82782 (18.8%) | 25384 (10.9%) | 57398 (27.5%) | 8374 (32.0%) | 4199 (27.4%) |
| Blood loss anemia | 6044 (1.4%) | 1699 (0.7%) | 4345 (2.1%) | 698 (2.7%) | 258 (1.7%) |
| Deficiency anemia | 26437 (6.0%) | 8626 (3.7%) | 17811 (8.5%) | 2401 (9.2%) | 1384 (9.0%) |
| Alcohol abuse | 34542 (7.8%) | 12501 (5.4%) | 22041 (10.5%) | 2206 (8.4%) | 3731 (24.4%) |
| Drug abuse | 29648 (6.7%) | 11538 (5.0%) | 18110 (8.7%) | 1480 (5.7%) | 3036 (19.8%) |
| Psychoses | 12536 (2.8%) | 4766 (2.1%) | 7770 (3.7%) | 602 (2.3%) | 1185 (7.7%) |
| Depression | 72698 (16.5%) | 27630 (11.9%) | 45068 (21.6%) | 4721 (18.1%) | 4192 (27.4%) |
| *Information collected during ED stay* | | | | | |
| Temperature (Celsius) | 36.76 (0.37) | 36.72 (0.32) | 36.80 (0.42) | 36.85 (0.61) | 36.73 (0.37) |
| Heart rate (bpm) | 78.14 (14.38) | 76.25 (12.84) | 80.24 (15.65) | 87.49 (20.13) | 79.97 (13.85) |
| Respiratory rate (bpm) | 17.25 (2.47) | 16.92 (1.87) | 17.60 (2.96) | 19.29 (4.55) | 17.03 (1.87) |
| Oxygen saturations (%) | 98.19 (2.94) | 98.55 (2.83) | 97.79 (3.01) | 97.58 (3.78) | 98.19 (2.90) |
| Systolic blood pressure (mmHg) | 127.39 (19.50) | 127.62 (18.56) | 127.13 (20.49) | 122.38 (22.22) | 128.72 (19.50) |
| Diastolic blood pressure (mmHg) | 73.56 (13.56) | 75.49 (12.68) | 71.42 (14.17) | 67.96 (15.13) | 75.97 (13.47) |
| Counts of medication prescription in the ED | 2.91 (3.30) | 1.79 (2.24) | 4.15 (3.81) | 5.33 (4.29) | 2.70 (3.21) |
| Counts of medication reconciliation | 6.11 (6.77) | 4.44 (5.88) | 7.96 (7.20) | 7.80 (7.53) | 5.17 (6.59) |
| ED length of stays (h) | 4.78 (7.47) | 0.30 (0.40) | 9.75 (8.41) | 5.62 (5.18) | 4.20 (7.83) |

The outcome statistics for the benchmark data are presented in Table 4, demonstrating a balanced stratification of the training and test data. In the overall cohort, 208,976 (47.34%) episodes require hospitalization, 26,145 (5.92%) episodes have critical outcomes, and 15,299 (3.47%) result in 72-hour ED reattendance.

**Table 4.** Outcome statistics of prediction tasks. The number of ED visits and their proportions in training and test data are shown for each outcome subgroup.

| | **Outcome** | | | | | Total (episodes) |
|---|---|---|---|---|---|---|
| | Hospitalization | ICU transfer in 12 hours | Inpatient mortality | Critical outcome | ED reattendance in 72 hours | |
| Training data | 167165 (47.34%) | 19791 (5.60%) | 3177 (0.90%) | 21026 (5.95%) | 12365 (3.50%) | 353150 (80%) |
| Test data | 41811 (47.36%) | 4816 (5.45%) | 773 (0.88%) | 5119 (5.80%) | 2934 (3.32%) | 88287 (20%) |
| Total (by outcome) | 208976 (47.34%) | 24607 (5.57%) | 3950 (0.89%) | 26145 (5.92%) | 15299 (3.47%) | 441437 (100%) |

### 3.2 Variable importance and ranking

With a descending order of variable importance extracted from RF, the top 10 variables selected for each benchmark task are presented in Table 5. Vital signs show significant predictive value in all three tasks. Age is also among the top predictive variables for all tasks, underscoring the impact of aging on emergency care utilization. While the triage level (i.e., ESI) is highly related to the hospitalization and critical outcome, it is not relevant to 72-hour ED reattendance. Conversely, despite its lower importance for hospitalization and critical outcomes, ED length of stay becomes the top variable for 72-hour ED reattendance prediction. The previous health utilization variable seems to be a less important feature for the ED-based tasks.

**Table 5.** Top 10 variables from each benchmark task by random forest variable importance

| **Hospitalization** | | **Critical outcomes** | | **72-hour ED reattendance** | |
|---|---|---|---|---|---|
| Variable | Importance | Variable | Importance | Variable | Importance |
| Age (years) | 0.1225 | Age (years) | 0.1008 | ED length of stays (hours) | 0.0843 |
| ESI at triage | 0.1122 | Systolic BP at triage (mmHg) | 0.0953 | Age (years) | 0.0843 |
| Systolic BP at triage (mmHg) | 0.0855 | Heart rate at triage (bpm) | 0.0935 | Systolic BP at ED (mmHg) | 0.0787 |
| Heart rate at triage (bpm) | 0.0846 | ESI at triage | 0.0847 | Diastolic BP at ED (mmHg) | 0.0762 |
| Diastolic BP at triage (mmHg) | 0.0816 | Diastolic BP at triage (mmHg) | 0.0835 | Heart rate at ED (bpm) | 0.0761 |
| Temperature at triage (Celsius) | 0.078 | Temperature at triage (Celsius) | 0.0757 | Temperature at ED (Celsius) | 0.0666 |
| Pain scale at triage | 0.0506 | Oxygen saturations at triage (%) | 0.0638 | Counts of medication reconciliation | 0.0506 |
| Oxygen saturations at triage (%) | 0.0496 | Respiratory rate at triage (bpm) | 0.0549 | Pain scale at triage | 0.0439 |
| Respiratory rate at triage (bpm) | 0.0403 | Pain scale at triage | 0.0468 | Oxygen saturations at ED (%) | 0.0399 |
| Hospitalizations in the past year | 0.0266 | Hospitalizations in the past year | 0.019 | Counts of medication reconciliation | 0.0398 |

BP: Blood Pressure

ED: Emergency Department

ESI: Emergency Severity Index

### 3.3 Benchmark task evaluation

Machine learning exhibited a higher degree of discrimination in predicting all three outcomes. Gradient boosting achieved an AUC of 0.881 (95% CI: 0.877-0.886) for the critical outcome and an AUC of 0.820 (95% CI: 0.818-0.823) for the hospitalization outcome. However, the corresponding performance for 72-hour ED reattendance was considerably lower. Compared with gradient boosting, deep learning could not achieve even higher performance. While traditional scoring systems did not show good discriminatory performance, interpretable machine learning-based AutoScore achieved an AUC of 0.846 (95% CI: 0.842-0.851) for critical outcomes with seven variables, and 0.793 (95% CI: 0.791-0.797) for hospitalization outcomes with 10 variables. Supplementary eTable 1 presents the performance of critical outcome prediction at ED disposition. Moreover, as shown in Table 6 and Figure 3, the performance of a variety of widely used machine learning and scoring systems is assessed by various metrics on the test set.

**Table 6:** Comparison of the performance of different models based on three different outcomes.

| **Hospitalization prediction at ED triage** | | | | | | | |
|---|---|---|---|---|---|---|---|
| Model | Threshold | AUROC (95% CI) | AUPRC (95% CI) | Sensitivity (95% CI) | Specificity (95% CI) | Runtime* | Number of variables |
| LR | 0.445 | 0.809 (0.807-0.812) | 0.776 (0.771-0.78) | 0.747 (0.735-0.752) | 0.725 (0.718-0.738) | 5 | 64 |
| RF | 0.489 | 0.819 (0.818-0.822) | 0.786 (0.784-0.789) | 0.754 (0.736-0.757) | 0.734 (0.731-0.751) | 58 | 64 |
| GB | 0.484 | 0.820 (0.818-0.823) | 0.794 (0.791-0.798) | 0.743 (0.740-0.76) | 0.743 (0.725-0.749) | 62 | 64 |
| MLP | 0.455 | 0.823 (0.822-0.826) | 0.797 (0.794-0.800) | 0.759 (0.754-0.763) | 0.735 (0.732-0.741) | 62 | 64 |
| ESI | 2 | 0.711 (0.709-0.714) | 0.632 (0.628-0.636) | 0.582 (0.578-0.586) | 0.784 (0.781-0.787) | 0 | 1 |
| AutoScore | 45 | 0.793 (0.791-0.797) | 0.756 (0.753-0.76) | 0.722 (0.717-0.749) | 0.721 (0.698-0.725) | 170 | 10 |
| NEWS | 1 | 0.581 (0.579-0.584) | 0.555 (0.552-0.559) | 0.565 (0.561-0.57) | 0.540 (0.537-0.544) | 0 | 6 |
| NEWS2 | 1 | 0.563 (0.56-0.566) | 0.538 (0.534-0.541) | 0.519 (0.514-0.522) | 0.563 (0.559-0.567) | 0 | 6 |
| REMS | 3 | 0.672 (0.669-0.675) | 0.610 (0.605-0.613) | 0.714 (0.709-0.716) | 0.564 (0.559-0.568) | 0 | 6 |
| MEWS | 2 | 0.559 (0.557-0.562) | 0.522 (0.518-0.526) | 0.300 (0.296-0.302) | 0.810 (0.808-0.813) | 0 | 6 |
| CART | 4 | 0.675 (0.673-0.678) | 0.618 (0.615-0.622) | 0.702 (0.698-0.706) | 0.586 (0.582-0.592) | 0 | 4 |
| Med2Vec | 0.441 | 0.814 (0.813-0.817) | 0.782 (0.779-0.786) | 0.743 (0.739-0.754) | 0.734 (0.725-0.738) | 1063 | 64+ 7930[#] |
| **Critical outcomes prediction at ED triage** | | | | | | | |
| Model | Threshold | AUROC (95% CI) | AUPRC (95% CI) | Sensitivity (95% CI) | Specificity (95% CI) | Runtime | Number of variables |
| LR | 0.065 | 0.863 (0.859-0.868) | 0.321 (0.308-0.336) | 0.782 (0.773-0.805) | 0.786 (0.760-0.796) | 7 | 64 |
| RF | 0.073 | 0.873 (0.867-0.878) | 0.377 (0.365-0.389) | 0.797 (0.773-0.803) | 0.792 (0.791-0.818) | 65 | 64 |
| GB | 0.065 | 0.881 (0.877-0.886) | 0.388 (0.374-0.405) | 0.801 (0.792-0.808) | 0.799 (0.796-0.807) | 76 | 64 |
| MLP | 0.05 | 0.883 (0.880-0.888) | 0.386 (0.375-0.404) | 0.810 (0.794-0.817) | 0.796 (0.794-0.815) | 376 | 64 |
| ESI | 2 | 0.804 (0.801-0.809) | 0.194 (0.187-0.205) | 0.870 (0.863-0.875) | 0.640 (0.637-0.643) | 0 | 1 |
| AutoScore | 66 | 0.846 (0.842-0.851) | 0.278 (0.267-0.293) | 0.804 (0.784-0.810) | 0.728 (0.726-0.747) | 166 | 7 |
| NEWS | 2 | 0.634 (0.627-0.64) | 0.141 (0.132-0.144) | 0.464 (0.453-0.472) | 0.795 (0.793-0.798) | 0 | 6 |
| NEWS2 | 2 | 0.616 (0.608-0.623) | 0.128 (0.122-0.131) | 0.410 (0.399-0.586) | 0.823 (0.531-0.824) | 0 | 6 |
| REMS | 5 | 0.686 (0.679-0.691) | 0.105 (0.102-0.111) | 0.681 (0.668-0.687) | 0.616 (0.613-0.619) | 0 | 6 |
| MEWS | 2 | 0.613 (0.606-0.618) | 0.103 (0.100-0.108) | 0.430 (0.417-0.439) | 0.770 (0.768-0.772) | 0 | 6 |
| CART | 6 | 0.707 (0.701-0.713) | 0.141 (0.132-0.148) | 0.590 (0.578-0.598) | 0.731 (0.728-0.733) | 0 | 4 |
| Med2Vec | 0.005 | 0.857 (0.853-0.863) | 0.342 (0.332-0.351) | 0.793 (0.775-0.801) | 0.770 (0.770-0.787) | 1063 | 64+ 7930# |
| **72-hour ED reattendance prediction at ED disposition** | | | | | | | |
| Model | Threshold | AUROC (95% CI) | AUPRC (95% CI) | Sensitivity (95% CI) | Specificity (95% CI) | Runtime | Number of variables |
| LR | 0.04 | 0.683 (0.677-0.697) | 0.155 (0.141-0.169) | 0.620 (0.604-0.643) | 0.642 (0.622-0.665) | 3 | 67 |
| RF | 0.05 | 0.662 (0.647-0.674) | 0.144 (0.132-0.157) | 0.602 (0.573-0.619) | 0.622 (0.617-0.625) | 28 | 67 |
| GB | 0.038 | 0.699 (0.689-0.712) | 0.162 (0.149-0.177) | 0.653 (0.618-0.673) | 0.631 (0.618-0.661) | 30 | 67 |
| MLP | 0.04 | 0.696 (0.687-0.709) | 0.160 (0.146-0.174) | 0.625 (0.602-0.675) | 0.652 (0.610-0.681) | 93 | 67 |
| LSTM | 0.038 | 0.697 (0.683-0.712) | 0.158 (0.144-0.171) | 0.637 (0.612-0.659) | 0.657 (0.651-0.678) | 10454 | 67^ |
| AutoScore | 27 | 0.673 (0.665-0.684) | 0.114 (0.107-0.124) | 0.621 (0.596-0.637) | 0.628 (0.622-0.665) | 180 | 12 |
| Med2Vec | 0.002 | 0.678 (0.670-0.694) | 0.139 (0.129-0.151) | 0.622 (0.570-0.640) | 0.630 (0.620-0.700) | 1063 | 64+ 7930# |

AUROC: The area under the receiver operating characteristic
AUPRC: The area under the precision-recall curve
CART: Cardiac Arrest Risk Triage

CI: Confidence interval
ESI: Emergency Severity Index
GB: Gradient Boosting
LSTM: Long short-term memory
LR: Logistic Regression
MEWS: Modified Early Warning Score
MLP: Multilayer Perceptron
NEWS: National Early Warning Score
NEWS2: National Early Warning Score 2
REMS: Rapid Emergency Medicine Score
RF: Random Forest
* The unit of the running time in seconds.
^ Include 7 temporal variables
# The dataset contains 7930 distinct ICD codes

**Figure 3**: Bar plots comparing the performance of various prediction models based on three different outcomes.

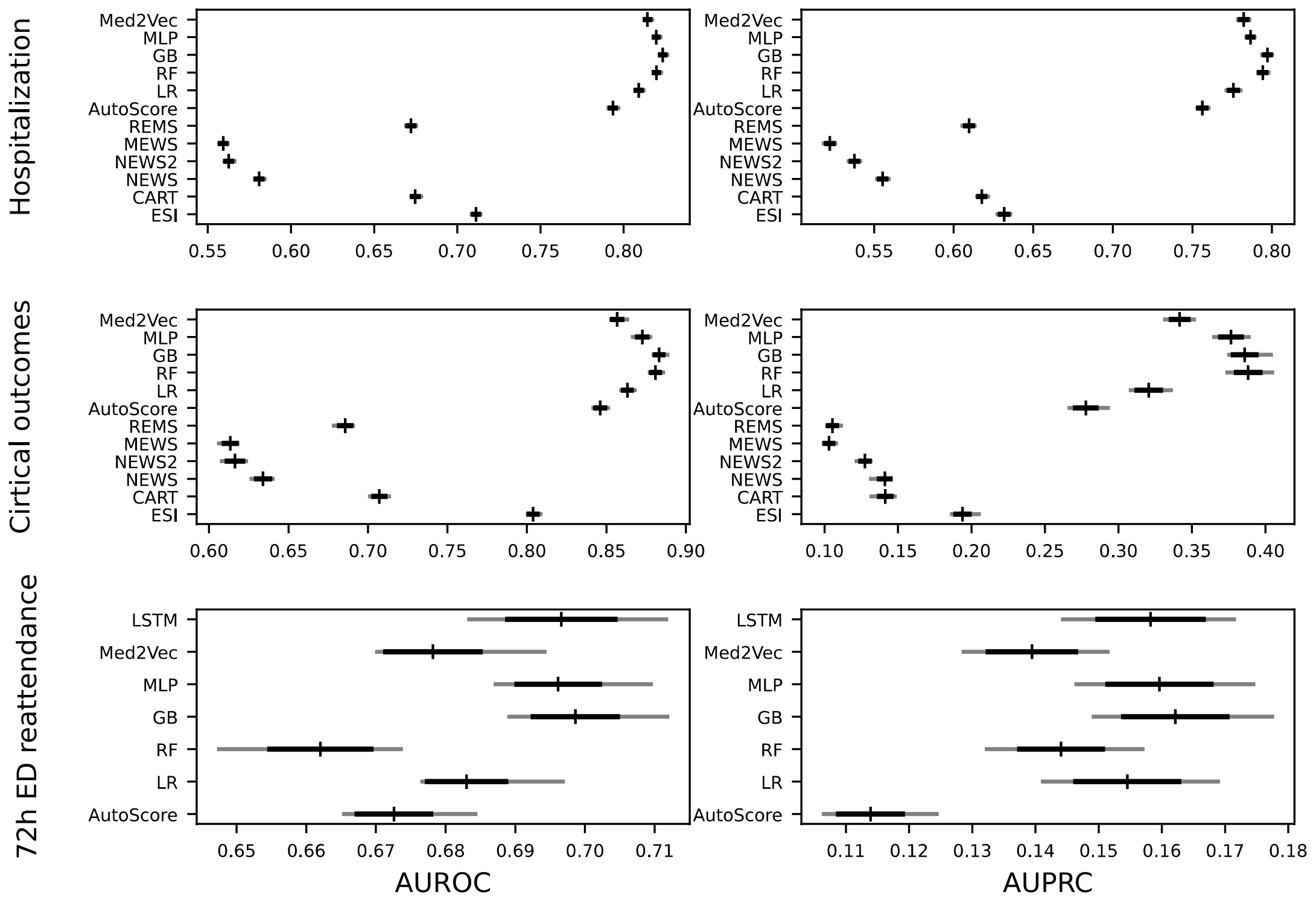

AUROC: The area under the receiver operating characteristic curve

AUPRC: The area under the precision-recall curve

CART: Cardiac Arrest Risk Triage

ESI: Emergency Severity Index

GB: Gradient Boosting

LSTM: Long short-term memory

LR: Logistic Regression

MEWS: Modified Early Warning Score

NEWS: National Early Warning Score

MLP: Multilayer Perceptron

NEWS2: National Early Warning Score, Version 2

REMS: Rapid Emergency Medicine Score

RF: Random Forest

# 4 Discussion

This paper proposed standardized benchmarks for future researchers interested in analyzing large-scale ED clinical data. Our study provides a pipeline to process raw data from the newly published MIMIC-IV-ED database, and generates a benchmark dataset, the first of its kind in the ED context. The benchmark dataset contains approximately half a million ED visits, and is conveniently accessible by researchers who plan to replicate our experiments or further build upon our work. Additionally, we demonstrated several triage prediction models (e.g., machine learning and clinical scoring systems) on routinely available information using this benchmark dataset for three ED-relevant outcomes: hospitalization, critical outcome, and ED reattendance. Our benchmark dataset also supports linkage to the main MIMIC-IV database, allowing researchers to analyze a patient's clinical course from the time of ED presentation through the hospital stay.

Our study showed that machine learning models demonstrated higher predictive accuracy, consistent with the previous studies[9,17,60]. Complex deep learning[61] models such as Med2Vec and LSTM did not perform better than simpler models. These results suggest that overly complex models do not necessarily improve performance with relatively simple and low-dimensional data in the ED. Furthermore, predictions made by black-box machine learning have critical limitations in clinical practice[62,63], particularly for decision-making in emergency care. Although machine learning models outperform in terms of predictive accuracy, the lack of explanation makes it challenging for frontline physicians to understand how and why the model reaches a particular conclusion. In contrast, scoring systems combine just a few variables using simple arithmetic and have a more explicit clinical representation[57]. This transparency allows doctors to understand and trust model outputs more easily and contributes to the validity and acceptance of clinical scores in real-world settings[64,65]. In our experiments, predefined scoring systems were unable to achieve satisfactory accuracy. However, AutoScore-based data-driven scoring systems complemented them with much higher accuracy while maintaining the advantages of the point-based scores[7].

The primary goals of ED triage prediction models are to identify high-risk patients accurately and to allocate limited resources efficiently. While physicians can generally determine the severity of a patient's acute condition, their decisions are often subjective and depend on an individual's knowledge and experience. This study explored data-driven methods to provide an objective assessment for three ED-relevant risk triaging tasks based on large-scale public EHRs. The open nature of the models makes them suitable for reproducibility and improvement. The scientific research community can make full use of the data and the triage prediction models to improve emergency care. In addition, the three ED triaging tasks are interrelated, yet represent distinct groups of predictors. Hospitalization and critical outcomes share a similar set of predictors, whereas the prediction of ED reattendances depend on various other variables.

This study contributes to the scientific community by standardizing research workflows and reducing barriers of entry for both clinicians and data scientists engaged in ED research. In the future, researchers may use this data pipeline to process raw MIMIC-IV-ED data. They may also develop new models and evaluate them against our ED-based benchmark tasks and prediction models. Additionally, our pipeline does not focus exclusively on ED data; we also provide linkages to the MIMIC-IV main database with all ICU and inpatient episodes. Data scientists interested in extracting ED data as additional variables and linking them to the other settings of the MIMIC-IV database can exploit our framework to streamline their research without consulting different ED physicians. With the help of this first large-scale public ED benchmark dataset and data processing pipeline, researchers can conduct high-quality ED research without needing a high level of technical proficiency.

This study has several limitations. First, although the study is based on an extensive database, it is still a single-center study. The performance of different methods used in this study may differ in other healthcare settings. Despite this, the proposed benchmarking pipeline could still be used as a reference for future big data research in the ED. Furthermore, examining whether models trained on the benchmark data generalize to other clinical datasets would be interesting. Second, the benchmark dataset established in this study is based on EHR data with routinely collected variables, where certain potential risk factors, such as socioeconomic status and neurological features, were not recorded. In addition, the dataset lacks sufficient information to detect out-of-hospital deaths, which may bias our models. Despite these limitations, the data processing pipeline can be leveraged widely when new researchers wish to conduct ED research using the MIMIC-IV-ED database.

## Supplementary Materials

**eTable1** Comparison of performance of different models applied to critical outcomes at ED disposition.

| **Critical Outcomes prediction at ED disposition** | | | | | | | |
|---|---|---|---|---|---|---|---|
| Model | Threshold | AUROC (95% CI) | AUPRC (95% CI) | Sensitivity (95% CI) | Specificity (95% CI) | Runtime* | Number of variables |
| LR | 0.059 | 0.857 (0.854-0.861) | 0.346 (0.333-0.359) | 0.775 (0.765-0.784) | 0.780 (0.769-0.784) | 8 | 67 |
| RF | 0.09 | 0.932 (0.930-0.935) | 0.567 (0.560-0.577) | 0.876 (0.852-0.879) | 0.831 (0.83-0.856) | 52 | 67 |
| GB | 0.077 | 0.934 (0.932-0.936) | 0.554 (0.546-0.565) | 0.856 (0.851-0.872) | 0.850 (0.837-0.852) | 79 | 67 |
| MLP | 0.059 | 0.937 (0.936-0.940) | 0.557 (0.551-0.566) | 0.878 (0.864-0.887) | 0.839 (0.837-0.849) | 79 | 67 |
| LSTM | 0.051 | 0.945 (0.943-0.947) | 0.59 (0.586-0.603) | 0.876 (0.866-0.892) | 0.860 (0.846-0.869) | 79 | 200 |
| ESI | 2 | 0.804 (0.799-0.808) | 0.194 (0.187-0.201) | 0.870 (0.861-0.875) | 0.640 (0.637-0.643) | 0 | 1 |
| NEWS | 2 | 0.634 (0.627-0.640) | 0.141 (0.132-0.147) | 0.464 (0.452-0.476) | 0.795 (0.793-0.798) | 0 | 6 |
| NEWS2 | 2 | 0.616 (0.610-0.623) | 0.128 (0.121-0.135) | 0.410 (0.399-0.592) | 0.823 (0.532-0.825) | 0 | 6 |
| REMS | 5 | 0.686 (0.682-0.691) | 0.105 (0.101-0.11) | 0.681 (0.671-0.689) | 0.616 (0.613-0.618) | 0 | 6 |
| MEWS | 2 | 0.613 (0.608-0.619) | 0.103 (0.099-0.109) | 0.430 (0.419-0.438) | 0.770 (0.768-0.772) | 0 | 6 |
| CART | 6 | 0.707 (0.701-0.713) | 0.141 (0.135-0.15) | 0.590 (0.576-0.598) | 0.731 (0.728-0.733) | 0 | 4 |

CART: Cardiac Arrest Risk Triage
ESI: Emergency Severity Index

GB: Gradient Boosting
LR: Logistic Regression
LSTM: Long short-term memory
MEWS: Modified Early Warning Score
MLP: Multilayer Perceptron
NEWS: National Early Warning Score
REMS: Rapid Emergency Medicine Score
RF: Random Forest
* The unit of the running time in seconds.

**eFigure1: Neural network structure of MLP, LSTM and Med2Vec.**

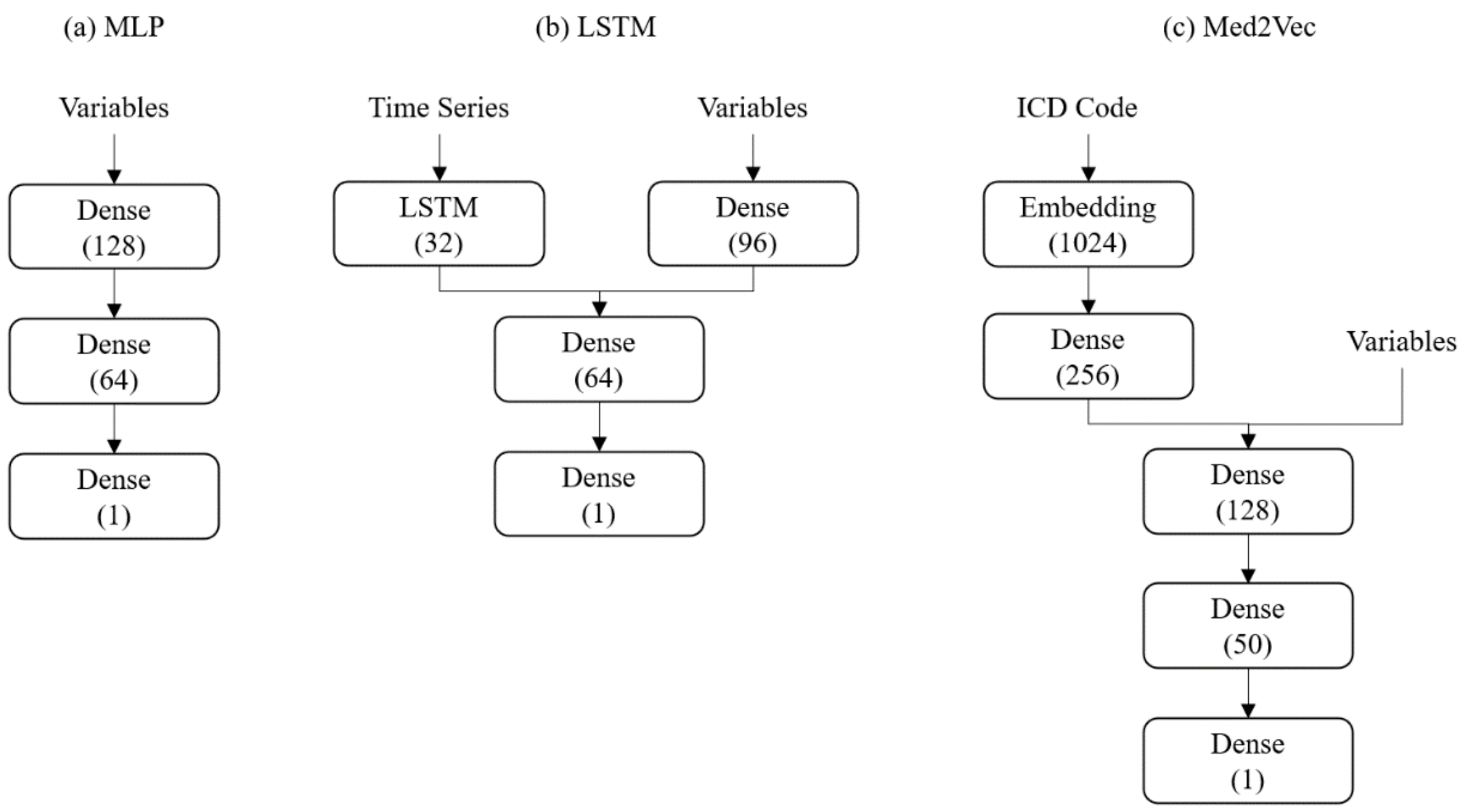

LSTM: Long short-term memory
MLP: Multilayer Perceptron